\documentclass[letterpaper]{article} 
\usepackage{aaai2026}  
\usepackage{times}  
\usepackage{helvet}  
\usepackage{courier}  
\usepackage[hyphens]{url}  
\usepackage{graphicx} 
\usepackage{natbib}  
\usepackage{caption} 
\usepackage{algorithm}
\usepackage{algorithmic}

\usepackage{colortbl}%
  \newcommand{\myrowcolour}{\rowcolor[gray]{0.925}}

\usepackage{tabularx}
\newcolumntype{C}{>{\centering\arraybackslash}X}
\usepackage{multirow}
\usepackage{amsmath}
\usepackage{booktabs}
\usepackage{amssymb}
\usepackage{xcolor}
\definecolor{linecolor1}{RGB}{244, 101, 98}
\definecolor{linecolor2}{RGB}{49, 133, 155}

\usepackage{newfloat}
\usepackage{listings}
\DeclareCaptionStyle{ruled}{labelfont=normalfont,labelsep=colon,strut=off} 
\floatstyle{ruled}
\newfloat{listing}{tb}{lst}{}
\floatname{listing}{Listing}
\title{Exploring Spectral Characteristics for Single Image Reflection Removal}
\author {
    Pengbo Guo\textsuperscript{\rm 1,2}, Chengxu Liu\textsuperscript{\rm 1}, Guoshuai Zhao\textsuperscript{\rm 1}, Xingsong Hou\textsuperscript{\rm 1}, Jialie Shen\textsuperscript{\rm 3}, Xueming Qian\textsuperscript{\rm 1}
}
\affiliations {
    \textsuperscript{\rm 1}Xi'an Jiaotong University\\
    \textsuperscript{\rm 2}Shanghai Innovation Institute\\
    \textsuperscript{\rm 3}City, University of London\\
}

\begin{document}

\maketitle

\begin{abstract}
Eliminating reflections caused by incident light interacting with reflective medium remains an ill-posed problem in the image restoration area. The primary challenge arises from the overlapping of reflection and transmission components in the captured images, which complicates the task of accurately distinguishing and recovering the clean background. Existing approaches typically address reflection removal solely in the image domain, ignoring the spectral property variations of reflected light, which hinders their ability to effectively discern reflections. In this paper, we start with a new perspective on spectral learning, and propose the Spectral Codebook to reconstruct the optical spectrum of the reflection image. The reflections can be effectively distinguished by perceiving the wavelength differences between different light sources in the spectrum. To leverage the reconstructed spectrum, we design two spectral prior refinement modules to re-distribute pixels in the spatial dimension and adaptively enhance the spectral differences along the wavelength dimension. Furthermore, we present the Spectrum-Aware Transformer to jointly recover the transmitted content in spectral and pixel domains. Experimental results on three different reflection benchmarks demonstrate the superiority and generalization ability of our method compared to state-of-the-art models.
\end{abstract}

\section{Introduction}
\label{sec:intro}

\begin{figure*}[tb]
    \centering
    \includegraphics[width=\textwidth]{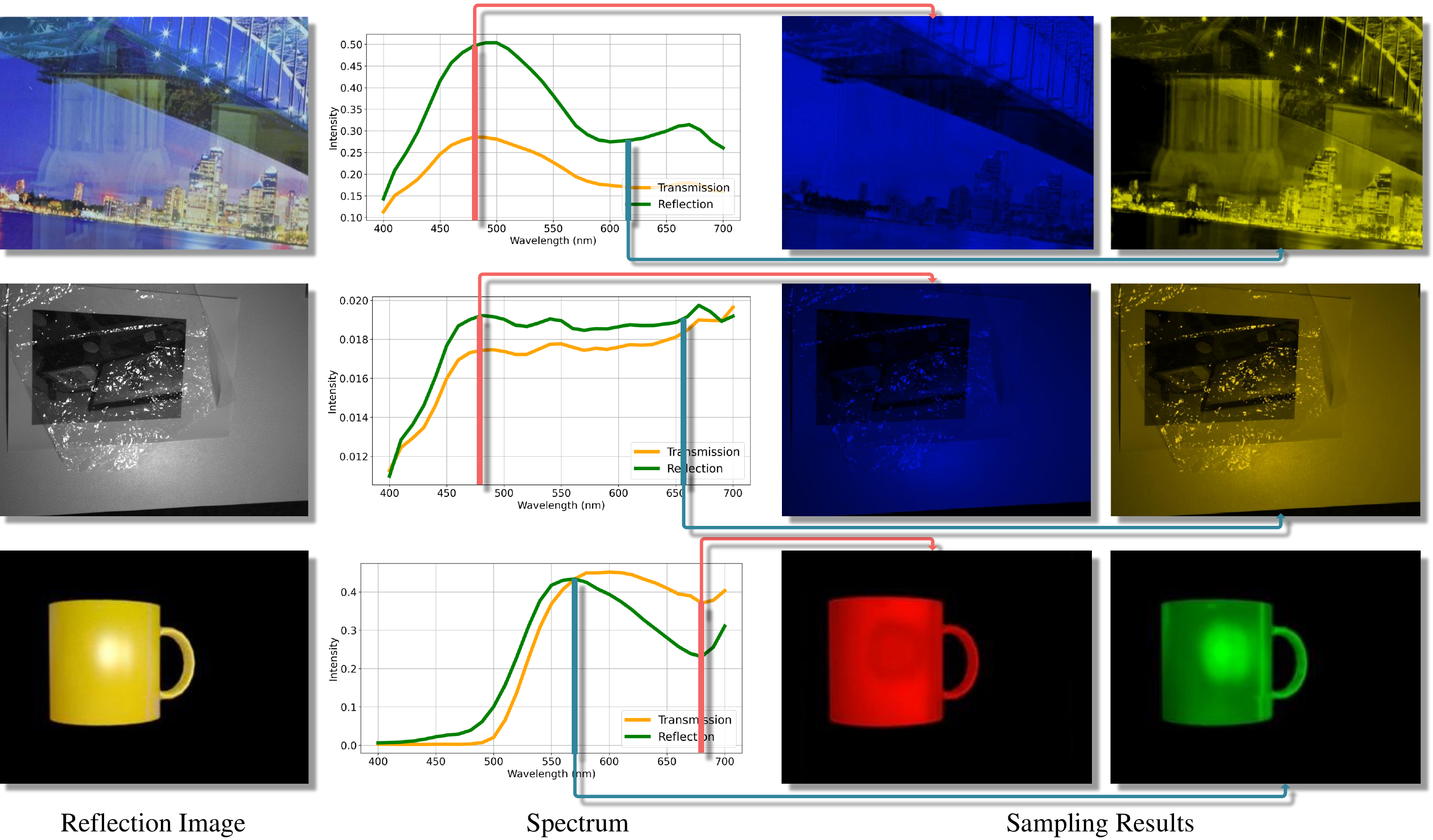}
    \caption{We use the latest SOTA spectral reconstruction method MST++~\cite{cai2022mst++} to estimate the spectrum from the given reflection and transmission images across various scenarios. By sampling at the spectral bands where intensity differences are larger, we can obtain the results nearly without reflections. Such observation demonstrates the necessity of leveraging the spectrum to guide the reflection removal process. The \textcolor{linecolor1}{red} lines represent sampling at the maximum intensity difference (without reflection), and the \textcolor{linecolor2}{blue} lines represent sampling at the minimum difference (with full reflection).}
    \label{fig:ref_change}
\end{figure*}

As a ubiquitous physical phenomenon in daily life, reflection is typically induced by transparent or shiny material such as glass or metal, introducing undesirable artifacts that degrade image quality and disrupt subsequent downstream vision tasks~\cite{sinha2012image,piccinelli2024unidepth,wang2025vggt}. Consequently, reflection removal has always been a crucial task in low-level vision.

Existing methods primarily focus on specific scenarios, \textit{e.g.}, transparent reflection~\cite{hu2021trash, hu2023single, zhu2024revisiting}, plastic film reflection~\cite{tang2024learning}, or specular highlight~\cite{fu2021multi} to remove reflections through carefully designed hand-crafted priors or networks. However, these approaches suffer from the following challenges: 1) severe overfitting due to insufficient training data causes models to recognize reflection patterns as simple color mappings, mistakenly identifying highly saturated regions as reflection while overlooking those with lower saturation; 2) being one of the most prevalent physical phenomena in nature, reflection does not occur in isolation across different scenarios. The various reflection degradations are visual results of incident light interacting with different reflective media~\cite{born2013principles}. Treating them separately will decrease the model's robustness and incur additional costs. Therefore, exploring the differential representations of various light sources is essential for reflection removal.

According to the physical principles underlying reflection, the energy distribution of reflected light exhibits significant differences when incident light traverses a reflective medium~\cite{born2013principles,zheng2015illumination,nguyen2018multi,qiu2023effectiveness}. Motivated by this, we conduct experiments across three different reflection scenarios, leveraging the wavelength differences between reflected and transmitted light to observe variations in reflection regions. As shown in Fig.~\ref{fig:ref_change}, by sampling specific bands where spectral differences are pronounced, we can obtain clean spectral images devoid of reflection degradations. More evidence could be found in the supplementary materials.

Building upon this observation, we focus on the optical characteristics of reflected light, finding that \textit{spectral information can be used to amplify the differences between reflection and transmission components}, which can be effectively utilized to address the ill-posed light source separation problem. Therefore, we aim to guide reflection removal by learning the spectrum of input reflection image, leveraging wavelength differences to separate different components. Specifically, we employ paired RGB-spectral images to train our Spectral Codebook that models the spectral distribution across bands. During reflection removal, the spectrum can be directly reconstructed using the pre-trained Spectral Codebook without incurring additional costs.

\begin{figure*}[tb]
    \centering
    \includegraphics[width=\textwidth]{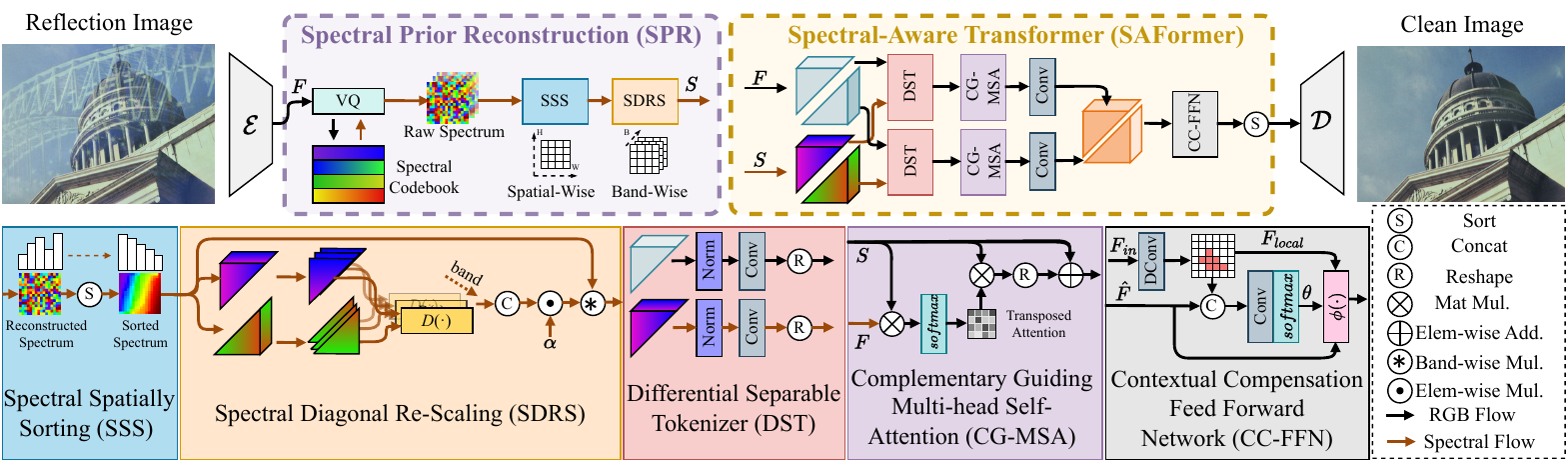}
    \caption{Overview of our proposed method. The Spectral Prior Reconstruction (SPR) module leverages the Spectral Codebook to reconstruct the spectrum of the input reflection image, and adaptively enhance its effectiveness. Our Spectral-Aware Transformer (SAFormer) differentially learns the cross-domain sorted features.}
    \label{fig:method}
\end{figure*}

To remove reflections embedded in captured images, we propose a novel single-image reflection removal network that leverages spectral guidance to discern differences between each component. The key insight of our method is to exploit the spectral characterization of reflection, eliminating reflection degradation and reconstructing the transmission content by jointly learning in both RGB and spectral domains. An overview is shown in Fig.~\ref{fig:method}, which comprises the Spectral Prior Reconstruction (SPR) module and the Spectral-Aware Transformer (SAFormer). Specifically, SPR first reconstructs the spectrum of the given reflection image, and refines it in both spatial and wavelength dimensions to amplify the differences between each component and eliminate redundant information. Subsequently, the Differential Separable Tokenizer (DST) separately encodes input features with different patterns, and the Complementary Guiding Multi-head Self-Attention (CG-MSA) collaboratively restores the clean background content. The CC-FFN is introduced to compensate for contextual information lost during differential learning.

Our contributions can be summarized as follows:
\begin{itemize}
\vspace{-0.1cm}
    \item We explore the intrinsic relationship between spectral wavelength differences and reflection imaging, highlighting the potential of using spectral bands with significant energy differences to effectively suppress reflections, offering a new insight for reflection removal.
    \item We propose the Spectral Codebook to reconstruct the spectrum of the input image without additional computational costs, and two spectral prior refinement modules to enhance its effectiveness in guiding reflection removal.
    \item We design a novel Spectrum-Aware Transformer, pioneering the integration of spectral learning into reflection removal task. By jointly learning in both RGB and spectral domains, our method effectively eliminates reflections and reconstructs the clean transmission content.
\end{itemize}

\section {Related Work}

\subsection{Single Image Reflection Removal}

Fewer constraints on data make single image-based methods more popular, but also incur challenges due to the limited information. To address this, traditional methods~\cite{levin2007user,yang2019fast} rely on handcrafted priors, assuming distinct statistical distributions for the transmission and reflection layers. However, the reliance on such rigid assumptions limits their robustness, often leading to failure when encountering complex scenes that violate these priors.

To overcome these limitations, recent works has largely shifted to deep learning-based solutions. Hu and Guo \shortcite{hu2021trash} simultaneously restore the reflection and transmission components, using a ReLU rectifier to enhance interaction between the two branches. Furthermore, Hu and Guo~\shortcite{hu2023single} propose a learnable residual term that effectively captures residual information during the decomposition process. Zhong et al.~\shortcite{Zhong_2024_CVPR} introduce language descriptions for different components in the reflection image to guide the removal more efficiently. Zhu et al.~\shortcite{zhu2024revisiting} design a maximum filter to locate the reflection regions, then use positional guidance to remove reflections. However, most methods focus on the design of complex modules, ignoring exploring the fundamental characteristics to handle reflection.

\vspace{-0.2cm}
\subsection{Spectral Reconstruction}
Hyperspectral imagery (HSI) captures fine-grained spectral signatures that offer richer texture and structural details compared to conventional RGB images. However, traditional HSI acquisition, which typically relies on spectral scanning~\cite{huang2022spectral}, is often impractical due to prohibitive equipment costs and long capture times.

Recently, learning-based hyperspectral reconstruction from a single RGB image has emerged as a prominent research direction. Xiong et al.~\shortcite{xiong2017hscnn} propose the HSCNN framework that utilizes input RGB images and compressive measurements for spectral reconstruction. Cai et al.~\shortcite{cai2022mst++} apply a Transformer-based architecture to the task of spectral reconstruction. In this work, we model the distribution of reflections in the spectral domain, leveraging spectral information to guide the reflection removal process.

\section{Our Approach}

As shown in Fig.~\ref{fig:method}, given a reflection image through the encoder, the Spectral Prior Reconstruction (SPR) module first leverages the Spectral Codebook to reconstruct its spectrum, and adaptively refine the spectrum along the spatial and wavelength dimensions, respectively. The input RGB features and refined spectrum are subsequently fed into the Spectrum-Aware Transformer (SAFormer) to jointly remove the reflection and restore the transmission content.

\begin{figure}[tb]
    \centering
    \includegraphics[width=0.48\textwidth]{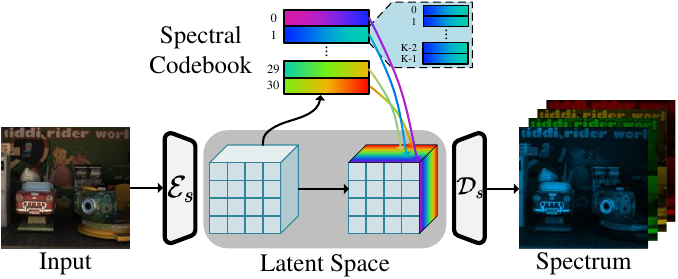}
    \caption{Illustration of training our Spectral Codebook on paired RGB-spectrum images.}
    \label{fig:dic}
\end{figure}

\subsection{Spectral Prior Reconstruction}
\label{SPR}
The wavelength variability in Fig.~\ref{fig:ref_change} inspires us to exploit spectral information to eliminate degradation caused by reflections. Unfortunately, directly reconstructing the spectrum presents significant challenges in properly guiding the restoration process and would require an additional reconstruction network with huge parameters. To address this, we propose the Spectral Codebook as a more efficient approach to reconstruct the reflection spectrum.

To better leverage the guidance of our reconstructed raw spectrum, the SSS and SDRS are proposed to optimize it along the spatial and wavelength dimensions, respectively. The former organizes pixels based on spectral intensity, enabling pixels with similar patterns are also adjacent in space. The latter adaptively re-scales the differences across bands, helping the network to better model the differential representations between reflection and transmission.

\paragraph{Spectral Codebook.}
We pre-train our Spectral Codebook on paired RGB-spectrum images for spectral reconstruction, specific dataset information can be obtained in experiments. To align with the discrete spectral data, we divide the entire codebook into 31 equal parts, each corresponding to a visible spectral band (\textit{e.g.}, 400-410nm, 410-420nm, etc.)~\cite{kaya2019towards}. As shown in Fig.~\ref{fig:dic}, given an RGB image $\boldsymbol{x} \in \mathbb{R}^{3 \times H \times W}$, it is first fed into the encoder to extract features $\hat{z}=\mathcal{E}_s(\boldsymbol{x}) \in \mathbb{R}^{n_{z} \times H \times W}$, where $H$, $W$, and $n_{z}$ represent the feature's height, width, and dimension of latent vector, respectively. Next, for the $i$-th band, the nearest neighbors $z^i_k \in \mathbb{R}^{n_{z}}$ of each pixel in $\hat{z}$ are found in the corresponding Spectral Codebook part $\mathcal{Z}^i \in \mathbb{R}^{K \times n_{z}}$, $K$ is the length of $i$-th codebook part, yielding the Band-Wise Vector-Quantized (Band-Wise VQ) representation $\tilde{z}$:
\begin{equation}
\begin{split}
     \tilde{z}_{h w}^i&=\mathbf{q}(\hat{z}_{h w})=\mathop{\arg\min}_{z_{k}^i \in \mathcal{Z}^i}(\|\hat{z}_{h w}-z_{k}^i\|_{2}),\\
     \tilde{z}&  = Concat(\tilde{z}^i), i \in \{0, 1, ..., 30\},
\end{split}
\end{equation}
where $\hat{z}_{h w}$ denotes each pixel in the spatial domain, $\mathbf{q}(\cdot)$ represents the element-wise quantization, and $Concat(\cdot)$ is the concatenation operation. The Decoder $\mathcal{D}_s$ maps the quantized representation $\tilde{z}$ back to the RGB space. The overall reconstruction paradigm can be formulated as:
\begin{equation}
    S=\mathcal{D}_s\left(z^{\prime}\right)=\mathcal{D}_s\left(\mathbf{q}\left(\mathcal{E}_s(x)\right)\right),
\end{equation}
where $S \in \mathbb{R}^{31 \times H \times W}$ denotes the reconstructed spectrum, $31$ is the number of spectral bands. During restoration, only the Spectral Codebook will be employed to reconstruct the spectrum according to the extracted RGB features.

The advantages of utilizing the Spectral Codebook are threefold: First, the spectral representations are spatially sparse while spectrally similar~\cite{cai2022mst++}, our Band-Wise VQ can better capture spatial differences while mitigating overemphasis on the redundant wavelength dimension. Second, the Spectral Codebook separately reconstructs each spectral band to align with the discrete distribution of the training data; at the same time, each part can recover the detailed spectral representation of its corresponding band. Third, our Spectral Codebook can be seamlessly integrated into the reflection removal network, enabling spectral reconstruction on the encoded features without introducing additional network components. Comparisons with other direct reconstruction approaches are provided in Tab.~\ref{tab:codebook}.

\paragraph{Spectral Spatially Sorting.}
Unlike common degradations such as noise or haze, reflection contains rich structural and texture details. Due to the inherent translational invariance of CNNs, when learning from local regions containing both reflection and transmission, the network often encounters unstable gradients, leading to restoration failures~\cite{sun2024restoring,cui2024revitalizing}. Therefore, we propose Spectral Spatially Sorting (SSS) to mitigate this issue without modifying the convolution paradigm, while also assisting subsequent modules in distinguishing different components.

Formally, given an input spectral features $S\in \mathbb{R}^{C\times H \times W}$, where $C$ denotes its channel, we perform sorting along the height $hi$ and width $wi$ dimensions as follows:
\begin{equation}
S^{\prime}, id = Sort_{wi}(Sort_{hi}(S)),
\end{equation}
where $Sort(\cdot)$ denotes the sorting operation, $id$ represents the sorting order, and $S^{\prime}$ indicates the output spectrum after sorting. The sorted spectrum is clustered into two regions in the top-left and bottom-right corners based on the spectral differences, thereby associating spatial distribution with pattern differences in the spectrum.

Notably, to align the spectral and RGB features for complementary learning, we also sort the input RGB features according to the order $id$. Before feeding the output into the decoder, we revert the output features to their original order. Such a design smoothes the gradients in local reflection regions, preventing blur or artifacts that could arise from gradient explosion during the restoration~\cite{dong2021location}.

\paragraph{Spectral Diagonal Re-Scaling.}

To enhance the differences between reflection and transmission components across wavelength dimension, \textit{i.e.,}, channel dimension for features, we introduce the Spectral Diagonal Re-Scaling (SDRS) module to adaptively re-scale the spectrum.

Specifically, we split the sorted spectral features $S^{\prime}$ spatially along the anti-diagonal line into two triangular regions: the top-left part and the bottom-right part, denoted as $S^{tl}$ and $S^{br}$, respectively. We then compute the spectral difference $D_i$ between these two regions for each band:
\begin{equation}
\begin{split}
 S^{tl}, & \ S^{br}= Split(S_i^{\prime}),\\
 D_{i}(S^{tl}, S^{br}) = &\left| \sum_{h=1}^{H_{tl}} \sum_{w=1}^{W_{tl}} S_{hw}^{tl} - \sum_{h=1}^{H_{br}} \sum_{w=1}^{W_{br}} S_{hw}^{br} \right|,\\
 D = Concat& (D_i), i \in \{0, 1, ..., B-1\},
\end{split}
\end{equation}
where $S_i$ denotes the $i$-th spectral band, $B$ represents the total number of bands, and $Split(\cdot)$ is the feature splitting operation. To ensure the adaptive re-scaling, we introduce a learnable scaling factor $\alpha$, which is applied element-wise multiplication $\odot$ with the computed spectral difference $D \in \mathbb{R}^{B \times 1}$, followed by a $softmax(\cdot)$ function to obtain the re-scaling weights $W \in \mathbb{R}^{B \times 1}$ for each band:
\begin{equation}
W = softmax(D \odot \alpha).
\end{equation}
Finally, we multiply the re-scaling weights $W$ with the input spectrum $S$ to obtain the refined spectral features $S_r$:
\begin{equation}
S_r = W \circledast S,
\end{equation}
where $\circledast$ denotes the channel-wise multiplication. The re-scaled spectrum exhibits greater differences along the wavelength dimension, mitigating redundancy and aiding models in precisely locating the reflection regions.

\subsection{Spectrum-Aware Transformer}

Thanks to the powerful long-range model ability, the Transformer-Based methods can model the full reflection structures for more complete removal. However, the self-attention mechanism struggles to adaptively relate features with different patterns over long distances~\cite{sun2024restoring}, limiting its ability to handle spatially varying reflections. Furthermore, improperly incorporating spectrum may result in the loss of color details from the original pixel domain.

To address these challenges, we employ a dual-domain learning transformer, enabling the optimized spectral prior to effectively guide restoration in the pixel domain. Specifically, our SAFormer comprises the Differential Separable Tokenizer (DST), which separately encodes features from different components, and the Complementary Guiding Multi-head Self-Attention (CG-MSA) to jointly remove reflection in both spectral and pixel domains. To compensate for the contextual information lost during sorting, we propose the Contextual Compensation Feed-Forward Network (CC-FFN) to make up for it.

\begin{table*}[!t]
\centering
\begin{tabularx}{\textwidth}{l l C C C C C C C C}
\toprule
\multirow{2}{*}{Methods}   & \multirow{2}{*}{Venue} & \multicolumn{2}{c}{$Nature$ (20)} & \multicolumn{2}{c}{$Real$(20)} & \multicolumn{2}{c}{$SIR^{2}$(454)} & \multicolumn{2}{c}{$Average$(494)} \\ 
\cmidrule(l{2pt}r{2pt}){3-4} \cmidrule(l{2pt}r{2pt}){5-6} \cmidrule(l{2pt}r{2pt}){7-8} \cmidrule(l{2pt}r{2pt}){9-10}
\myrowcolour
& & PSNR$\uparrow$ & SSIM$\uparrow$ & PSNR$\uparrow$ & SSIM$\uparrow$ & PSNR$\uparrow$ & SSIM$\uparrow$ & PSNR$\uparrow$& SSIM$\uparrow$ \\ \midrule
Input Image & - & 20.44 & 0.785 & 18.96 & 0.733 & 22.76 & 0.885 & 22.51 & 0.884 \\ \midrule
\myrowcolour
BDN &  ECCV 2018 & 18.83 &  0.738 &  18.64 &  0.726 & 21.61 & 0.854 & 21.50 & 0.844 \\
FRS & CVPR 2019 & 20.01 & 0.756 & 18.63 & 0.719 & 22.23 & 0.867 & 21.99 & 0.867 \\
\myrowcolour
Zhang et al. & CVPR 2018 & 22.31 & 0.804 & 20.16 & 0.767 & 23.07 & 0.869 & 22.92 & 0.862 \\
ERRNet & CVPR 2019 & 22.57 & 0.807 & 20.67 & 0.781 & 22.97 & 0.885 & 22.85 & 0.877 \\
\myrowcolour
RMNet & CVPR 2019 & 21.08 & 0.730 & 19.93 & 0.718 & 21.66 & 0.843 & 21.57 & 0.834 \\
Kim et al. & CVPR 2020 & 20.10 & 0.759 & 20.22 & 0.735 & 23.57 & 0.877 & 23.30 & 0.886 \\
\myrowcolour
IBCLN & CVPR 2020 & 23.90 & 0.787 & 21.42 & 0.769 & 24.05 & 0.888 & 23.94 & 0.878 \\
YTMT & NeurIPS 2021 & 20.69 & 0.777 & 22.94 & 0.815 & 23.57 & 0.889 & 23.43 & 0.882 \\
\myrowcolour
LANet & ICCV 2021 & 23.51 & {0.810} & {23.40} & \underline{0.826} & 23.04 &0.898 & 23.07 & 0.891 \\
PNACR & ACM MM 2023 & {23.93} & 0.807 & 22.57 & 0.806 & 24.14 & 0.894 & 24.06 &   0.888 \\
\myrowcolour
DSRNet & ICCV 2023 & 21.24 & 0.789 & \underline{24.23} & 0.820 & {25.45} & {0.901} & {24.65} & {0.893} \\
\midrule
$\text{Zhu et al.}^\ast$ & CVPR 2024 & \underline{25.96} & \underline{0.843} & 23.82 & 0.817 & 25.45 & 0.910 &  25.40 & 0.904 \\
\myrowcolour
$\text{Zhong et al.}^\dagger$ & CVPR 2024 & 23.87 & 0.812 & 24.05 & 0.824 & \textbf{25.86} & \textbf{0.919} & 25.72 & \textbf{0.914} \\
$\text{L-DiffER}^\dagger$ & ECCV 2024 & 23.95 & 0.831 & 23.77 & 0.821 & 25.18 & 0.911 & 25.08 & 0.905 \\
\midrule
\myrowcolour
Ours & - & \textbf{25.99} & \textbf{0.850} &\textbf{24.36} & \textbf{0.834} & \textbf{25.86} & \underline{0.912} &  \textbf{25.81} & \underline{0.907} \\
\bottomrule   
\end{tabularx}
\vspace{-1mm}
\caption{Quantitative comparison on three reflection datasets. \textbf{Bold} and \underline{underlining} indicate the best and the second best performance, respectively. $\ast$ denotes method using additional training data, and $\dagger$ represents using additional language labels.}
\label{tab:exp}
\end{table*}

\paragraph{Differential Separable Tokenizer.}
Vanilla Transformer considers only positional encoding during tokenization, causing both reflection and transmission components to be projected into the same token, which further aggravates the confusion during global attention. To address this problem, we propose the DST to separately extract singular pattern features without interference from other components.

In terms of formula, given the sorted input RGB features $F_r \in \mathbb{R}^{C \times H \times W}$ and the refined spectrum $S_r \in \mathbb{R}^{C \times H \times W}$, DST applies the $Split(\cdot)$ operation to divide the features along the anti-diagonal line into top-left part $F^{tl}$ and bottom-right part $F^{br}$, promoting that tokenization is performed on features with singular pattern,
\begin{equation}
\begin{split}
    F^{tl}, F^{br} &= Split(F_{r}),\\
    S^{tl}, S^{br} &= Split(S_{r}).\\
\end{split}
\end{equation}
For each part, DST aggregates the features in the channel dimension using a $1\times1$ convolution kernel. Afterward, we apply a $3\times3$ depth-wise convolution to encode the spatial-pattern-related context of the sorted features. Finally, we reshape the projections, yielding the final output tokens:
\begin{equation}
\begin{split}
    \mathcal{Q}^{tl}=W_\mathcal{Q}^{tl}S^{tl}, \mathcal{K}^{tl}&=W_\mathcal{K}^{tl}F^{tl}, \mathcal{V}^{tl}=W_\mathcal{V}^{tl}F^{tl},\\
    \mathcal{Q}^{br}=W_\mathcal{Q}^{br}S^{br}, \mathcal{K}^{br}&=W_\mathcal{K}^{br}F^{br}, \mathcal{V}^{br}=W_\mathcal{V}^{br}F^{br},\\
\end{split}
\end{equation}
where $W_\mathcal{Q}, W_\mathcal{K}, W_\mathcal{V}$ denote the projection convolutions of the queries, keys, and values, respectively. Through DST, we project the spectral features as the query $\mathcal{Q}\in \mathbb{R}^{m\times HW}$, with the RGB features as the key $\mathcal{K}\in \mathbb{R}^{m\times HW}$ and value $\mathcal{V}\in \mathbb{R}^{m\times HW}$, where $m$ denotes the dimensions of the tokens.

\paragraph{Complementary Guiding Multi-head Self-Attention.}
To ensure the spectral prior can correctly guide the restoration, and harmoniously fuse features from both domains, we propose the CG-MSA to complementary restoring the degraded reflection regions. Given the output tokens of DST $\mathcal{Q}, \mathcal{K}$, and $\mathcal{V}$, our CG-MSA performs matrix multiplication on them sequentially, yielding the final output:
\begin{equation} 
\begin{split}
    & \text{CG-MSA}(\mathcal{Q},\mathcal{K},\mathcal{V})=softmax(\frac{\mathcal{Q}\mathcal{K}^T}{\sqrt{m}})\mathcal{V}.
\end{split}
\end{equation}
To mitigate the rapid increase of similarity in the overlapping reflection regions after multiplication, we introduce $\sqrt{m}$ before the $softmax(\cdot)$ function to scale the inputs into a reasonable range. The overall computation of our CG-MSA can be formulated as follows,
\begin{equation}
\begin{split}
    F_{o}^{tl} =\text{CG-MSA} (\text{LN}&(\mathcal{Q}^{tl},\mathcal{K}^{tl},\mathcal{V}^{tl}))\oplus F^{tl},\\
    F_{o}^{br} =\text{CG-MSA} (\text{LN}&(\mathcal{Q}^{br},\mathcal{K}^{br},\mathcal{V}^{br}))\oplus F^{br},\\
    \hat{F} = F_{o}^{tl} &\oplus F_{o}^{br},
\end{split}
\end{equation}
where $F_{o}^{tl}\in \mathbb{R}^{C\times H \times W}$ and $F_{o}^{br}\in \mathbb{R}^{C\times H \times W}$ denote the separated RGB features, $\hat{F}\in \mathbb{R}^{C\times H \times W}$ represents the final output, and $\oplus$ denotes the element-wise addition. Since the features in both domains exhibit similar distributions, such a design facilitates complementary learning with clearer spectral guidance. Moreover, the varying sensitivity of different spectral bands to texture can further enrich local details.

\begin{equation}
\begin{split}
    F_{local}& = \text{DConv}(F_{in}),\\
    \theta = softmax&(\text{Conv}_{1 \times 1}(Concat(F_{local}, \hat{F}))),\\
    F_{out} = &\ \theta\hat{F} \oplus (1-\theta)F_{local}.
\end{split}
\end{equation}
In this way, CC-FFN effectively compensates for the context lost during sorted learning with little cost.

\begin{figure*}[tb]
    \centering
    \includegraphics[width=\textwidth]{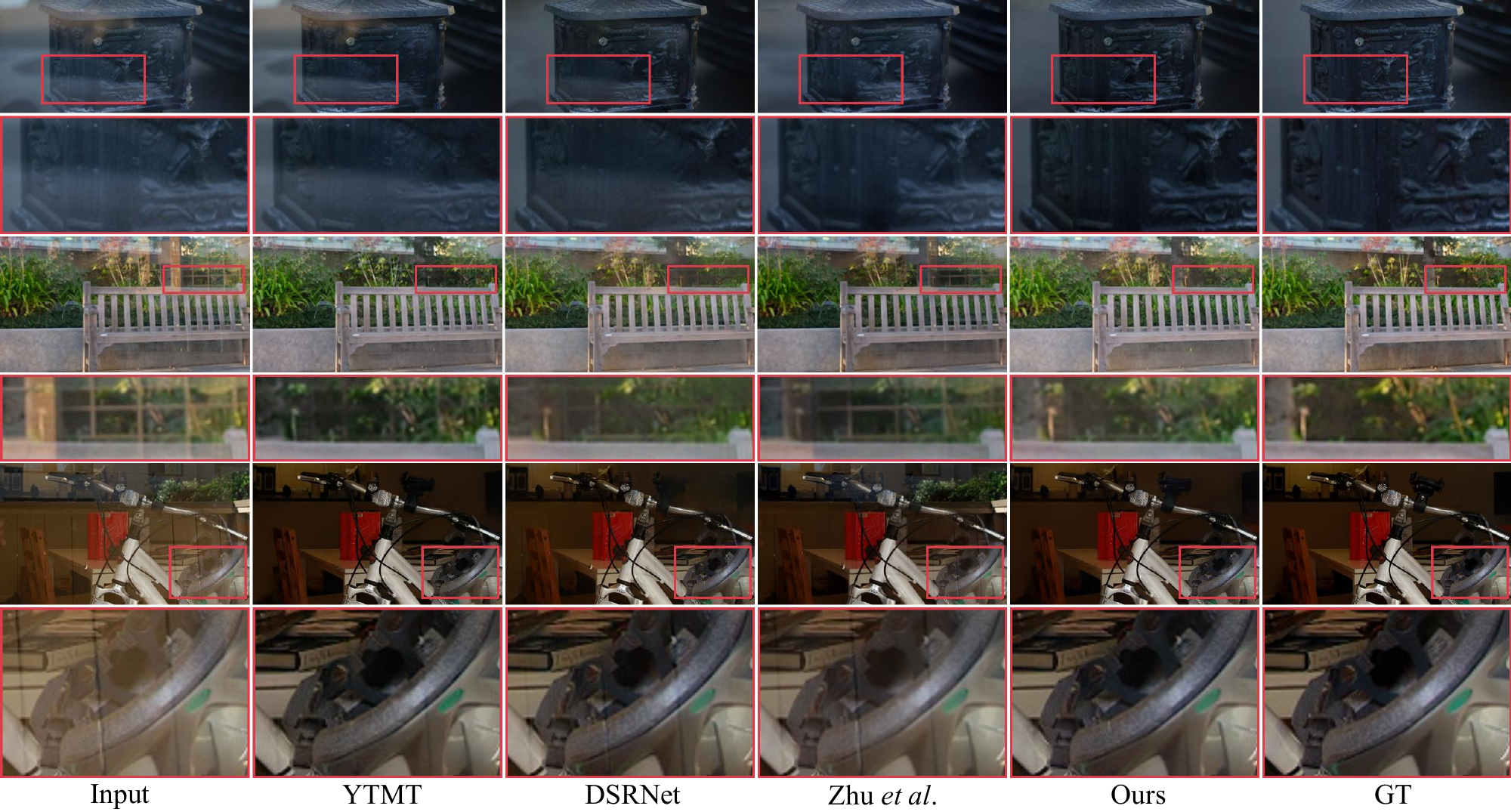}
    \caption{Visual results of the restored transmission images from the real-world reflection datasets~\cite{zhang2018single}.}
    \label{fig:result2}
\end{figure*}

\paragraph{Contextual Compensation Feed-Forward Network.}

While our method tackles the ill-posed overlapping components learning problem, it may cause the loss of contextual information. For example, if a part of an object is affected by reflection, the DST may separately project it into different tokens, thereby degrading the structural integrity. Therefore, we propose the Contextual Compensation Feed-Forward Network (CC-FFN), which interactively fuses features before and after sorting to compensate for the local context.

Specifically, as illustrated in Fig.~\ref{fig:method}, CC-FFN employs a deformable convolution layer~\cite{zhu2019deformable} $\text{DConv}(\cdot)$ to emphasize the local contextual features $F_{local}\in \mathbb{R}^{C\times H \times W}$ from the input features $F_{in}\in \mathbb{R}^{C\times H \times W}$. Then, CC-FFN~concatenates the local features $F_{local}$ with the output features $\hat{F}$ from CG-MSA, and uses a $1\times1$ convolution layer $\text{Conv}_{1 \times 1}(\cdot)$ to estimate the feature fusion ratio $\theta$ for weighted fusion of $F_{local}$ and $\hat{F}$,

\section{Experiments}
\label{sec.exp}

\subsection{Datasets}

We follow the existing settings for fairness~\cite{hu2023single,zhu2024revisiting}, using 90 real reflection pairs from Real dataset~\cite{zhang2018single}, 200 real pairs from Nature dataset~\cite{li2020single}, and 7,643 synthesized pairs from the PASCAL VOC dataset~\cite{everingham2010pascal}. For testing, we evaluate all methods on the Nature, Real, and $\text{SIR}^2$~\cite{wan2017benchmarking} datasets.

We further evaluate our methods in the other two reflection scenarios. The film reflection removal dataset~\cite{tang2024learning} comprising plastic film reflection images captured from different angles using polarized cameras. SSHR~\cite{fu2023towards} is a specular highlight removal dataset containing 117,000 training and 18,000 testing images.

\subsection{Implementation details}
\label{sec:expdetails}
To pre-train our spectral codebook, we use the NTIRE 2022 Spectral Reconstruction Dataset~\cite{arad2022ntire}, which contains 1,000 RGB-spectrum pairs. During training the removal network, we use the Adam optimizer~\cite{kingma2014adam} with $\beta_1$ = 0.9 and $\beta_2$ = 0.999. The learning rate is fixed $10^{-4}$.

\subsection{Comparisons with State-of-the-art Methods}
For fair comparisons, we follow the settings in DSRNet and compare with 14 SOTA methods. For the film reflection and specular highlight removal, we use the same settings in \cite{tang2024learning} and \cite{fu2023towards}. We obtain the performance from the latest SOTA work and their original papers.

\begin{table}[tb]\small
  \begin{minipage}[t]{0.24\textwidth}
    \centering
    \setlength{\tabcolsep}{1.9 mm}
    \begin{tabular}{lcc}
      \toprule
      Methods & PSNR & SSIM \\
      \midrule
      \myrowcolour
      SHIQ & 21.10 & 0.7465 \\
      Polar-HR & 22.18 & 0.7102 \\
      \myrowcolour
      Uformer & 31.85 & 0.9457 \\
      Restormer & 34.51 & 0.9759 \\
      \myrowcolour
      Tang et al. & \underline{36.94} & \textbf{0.9850} \\
      Ours & \textbf{37.15} & \underline{0.9826} \\
      \bottomrule
    \end{tabular}
    \vspace{-1.2mm}
    \captionof{table}{Quantitative comparison on film reflection dataset.}
    \label{tab:film}
  \end{minipage}
  \hspace{0.2mm}
  \begin{minipage}[t]{0.22\textwidth}
    \centering
    \setlength{\tabcolsep}{1.8 mm}
    \begin{tabular}{lcc}
      \toprule
      Methods & PSNR & SSIM \\
      \midrule
      \myrowcolour
      Yang & 23.243 & 0.894 \\
      Fu & 23.270 & 0.881 \\
      \myrowcolour
      JSHDR & 26.979 & 0.895 \\
      Wu & 25.731 & 0.894 \\
      \myrowcolour
      TSHR & \underline{28.633} & \textbf{0.940} \\
      Ours & \textbf{30.257} & \underline{0.936} \\
      \bottomrule
    \end{tabular}
    \vspace{-1.2mm}
    \captionof{table}{Quantitative comparison on SSHR dataset.}
    \label{tab:specular}
  \end{minipage}
\end{table}

\begin{figure*}[tb]
    \centering
    \begin{minipage}[t]{0.4\textwidth}
        \centering
        \small
        \setlength{\tabcolsep}{0.45 mm}
        \begin{tabular}{cccccccc}
        \toprule
        \multirow{2}{*}{Base} & \multicolumn{2}{c}{SPR} & \multicolumn{3}{c}{SAFormer} & \multirow{2}{*}{PSNR} & \multirow{2}{*}{SSIM} \\
        \cmidrule[0.1pt](lr{0.125em}){2-3}
        \cmidrule[0.1pt](lr{0.125em}){4-6}
        \myrowcolour%
           & SSS & SDRS & DST & CG-MSA &  CC-FFN &     &   \\
        \midrule
        \checkmark  &  & & & &          &  24.65  &  0.893 \\   
        \myrowcolour%
        \checkmark  & \checkmark& & & & & 24.99  &   0.902 \\   
        \checkmark & \checkmark& \checkmark& &  &   &25.11  &   0.902 \\   
        \myrowcolour%
        \checkmark  & & & \checkmark& \checkmark& \checkmark& 25.47  & 0.904  \\
        \checkmark  & \checkmark&  \checkmark& \checkmark&  & & 25.43  & 0.904 \\
        \myrowcolour%
        \checkmark & \checkmark&  \checkmark&  \checkmark&  \checkmark & & 25.75  &   0.906 \\
        \checkmark  &\checkmark&\checkmark&\checkmark&\checkmark& \checkmark& \textbf{25.81}  &  \textbf{0.907}   \\
        \bottomrule
        \end{tabular}
        \captionof{table}{Results of ablation study on our methods. SPR: Spectral Prior Reconstruction. SAFormer: Spectrum-Aware Transformer.}
        \label{tab:abl}
    \end{minipage}\hspace{1mm}
    \begin{minipage}[t]{0.26\textwidth}
        \centering
        \small
        \setlength{\tabcolsep}{1.4 mm}
        \renewcommand\arraystretch{0.95}
        \begin{tabular}{lcc}
        \toprule
        Methods    & PSNR  & SSIM   \\
        \midrule
        \myrowcolour
        RGB Stream &  24.65  & 0.893 \\
        Spectrum Stream &  23.81  & 0.886 \\
        \midrule
        \myrowcolour
        \textit{w/o} Spectrum &  24.65  & 0.893 \\
        Weighted Attn &  25.06  & 0.896 \\
        \myrowcolour
        Memory Bank & 25.16 &  0.902\\
        Sparse Coding &   25.19  &  0.898 \\
        \myrowcolour
        Vanilla VQ &   25.32  &  0.900 \\
        \midrule
        Ours & \textbf{25.81} & \textbf{0.907} \\
        \bottomrule
        \end{tabular}
        \captionof{table}{Ablation results on network architecture and spectral reconstruction strategies.}
        \label{tab:codebook}
    \end{minipage}
    \hspace{0.5mm}
    \begin{minipage}[t]{0.3\textwidth}
        \centering
        \small
        \vspace{-21.2mm}
        \includegraphics[width= \textwidth]{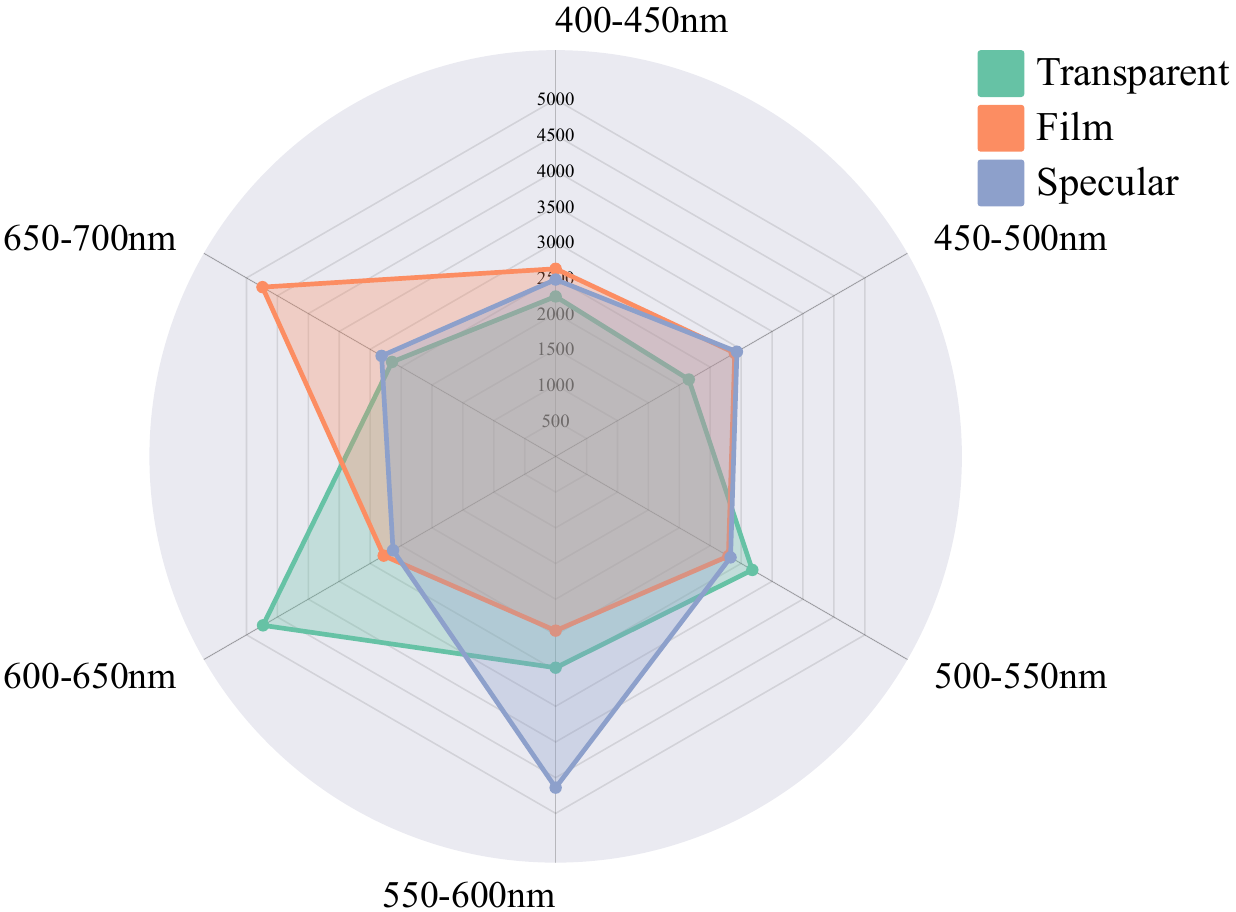}
        \caption{Visualization of code activation frequencies in different reflection scenarios.}
        \label{fig:codeact}
    \end{minipage}
\end{figure*}

\begin{figure}[tb]
    \centering
    \includegraphics[width=0.48 \textwidth]{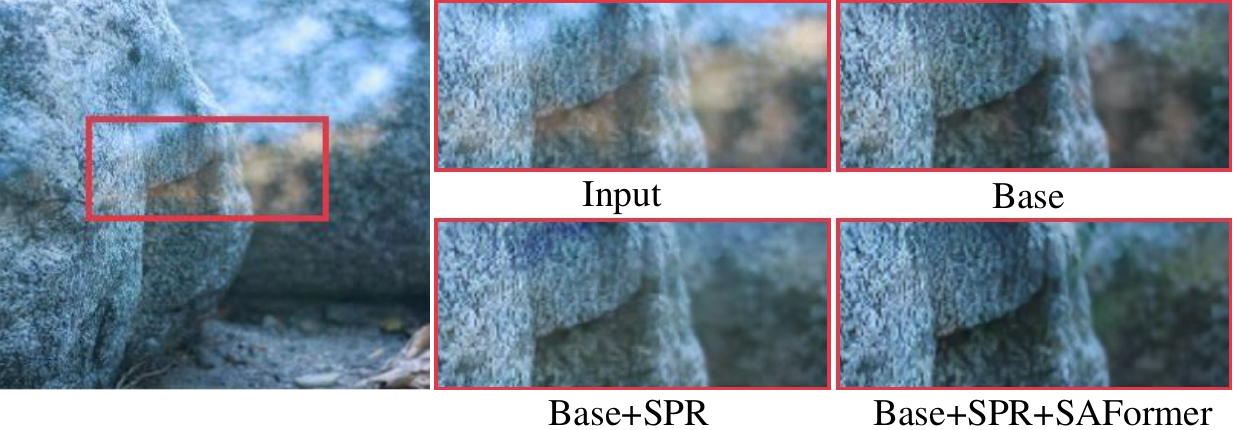}
    \vspace{-6mm}
    \caption{Visualization of ablation study on our method.}
    \label{fig:abl}
\end{figure}

\paragraph{Quantitative Comparison.}
Our method achieves SOTA performance on three reflection removal benchmarks in Tab.~\ref{tab:exp}. Such superiority stems from our novel spectral-pixel interactive learning, which effectively disentangles complex reflection artifacts. Furthermore, our model surpasses even recent methods that use additional data~\cite{zhu2024revisiting} or language priors~\cite{Zhong_2024_CVPR,LDiffER}. We further demonstrate the robustness and generalization of our approach on film reflection~\cite{tang2024learning} and specular highlight removal (SSHR)~\cite{fu2023towards} datasets in Tab.~\ref{tab:film} and \ref{tab:specular}, where it consistently outperforms existing methods.

\vspace{-0.1cm}
\paragraph{Qualitative Comparison.}
To further compare the visual qualities of different approaches, we present visual results restored by our method and other SOTA approaches on different reflection datasets in Fig.~\ref{fig:result2}. Our method delivers better visual quality, as evidenced by less reflective residue, correctly restoring the color in the reflection regions, and recovering more detailed textures. More visual results can be referred to the supplementary materials.

\subsection{Ablation study}
In this section, we conduct ablation on the proposed method and analyze the effect of the different network architectures. In addition, we further evaluate the reliability and robustness of our Spectral Codebook in different scenarios. More ablations could be obtained in the supplementary materials.

\paragraph{Individual Components.}
We analyze the effects of each component in our methods in Tab.~\ref{tab:abl} and Fig.~\ref{fig:abl}. We directly remove our method with the ResBlocks as the ``Base" model and progressively add each component for ablations. The introduction of SPR significantly improves PSNR from 24.65dB to 25.11dB, verifying its ability to eliminate reflections via spectral learning. Adding each component separately in the SAFormer all lead to favorable performance improvements, demonstrating the effectiveness of each component in our method. For visual results in Fig.~\ref{fig:abl}, the reflection degradations are well suppressed with the addition of SPR, proofing that the spectral prior can effectively help the model to distinguish reflection components. The introduction of SAFormer removes the residual reflections, further recovering the lost texture and color details.

\paragraph{Network Architecture.}
To evaluate the reliability of our SAFormer's dual-domain learning paradigm, we compare its performance with two single-stream networks in spectral and pixel domains, respectively. The results in Tab.~\ref{tab:codebook} confirm that the dual-stream design is crucial, significantly outperforming both baselines. This performance gain stems from the synergy between the two domains: the pixel stream preserves robust spatial structures, while the spectral stream provides the fine-grained texture necessary for restoration.

\paragraph{Spectral Codebook.}
To prove the superiority of our Spectral Codebook in estimating spectrum, we compare the performance of our Band-Wise VQ with other similar approaches in Tab.~\ref{tab:codebook}. It performs favorably against the other four approaches. This is because they either fail to provide spatially fine-grained reconstruction or are unable to separately model the spectral representations of different spectral bands. Therefore, we use the Band-Wise VQ with our Spectral Codebook to maximize reconstruction efficacy.

Fig.~\ref{fig:codeact} shows the activation frequencies of the codes in our Spectral Codebook. Significant distribution differences between different scenarios could be observed, which proves the robustness and sensitivity of our Spectral Codebook in learning diverse real-world reflections.

\section{Conclusion}

In this paper, we explore the spectral characteristics of reflection imaging, introducing a new perspective to handle reflections by learning the spectral difference between reflection and transmission components. In particular, we propose the Spectral Codebook to directly reconstruct the spectrum of input reflection image, and design two spectral refinement modules to enhance its efficacy. Furthermore, we propose the SAformer to jointly learn in both pixel and spectral domains. Such designs effectively eliminate reflections in various real-world scenarios and recover the detailed textures obscured by complex reflection degradations. Extensive experimental results show its advantages and the potential of spectral learning in the image restoration area.

\bibliography{main}

\end{document}